\documentclass[11pt,a4paper]{article}
\usepackage[hyperref]{eacl2021}
\usepackage{times}
\usepackage{latexsym}
\usepackage{graphicx}
\usepackage[hang]{footmisc}
\usepackage{listings}

\newcommand\YAMLcolonstyle{\color{red}\mdseries}
\newcommand\YAMLkeystyle{\color{black}\bfseries}
\newcommand\YAMLvaluestyle{\color{blue}\mdseries}

\makeatletter

\newcommand\language@yaml{yaml}

\expandafter\expandafter\expandafter\lstdefinelanguage
\expandafter{\language@yaml}
{
  keywords={true,false,null,y,n},
  keywordstyle=\color{darkgray}\bfseries,
  basicstyle=\YAMLkeystyle,                                 
  sensitive=false,
  comment=[l]{\#},
  morecomment=[s]{/*}{*/},
  commentstyle=\color{purple}\ttfamily,
  stringstyle=\YAMLvaluestyle\ttfamily,
  moredelim=[l][\color{orange}]{\&},
  moredelim=[l][\color{magenta}]{*},
  moredelim=**[il][\YAMLcolonstyle{:}\YAMLvaluestyle]{:},   
  morestring=[b]',
  morestring=[b]",
  literate =    {---}{{\ProcessThreeDashes}}3
                {>}{{\textcolor{red}\textgreater}}1     
                {|}{{\textcolor{red}\textbar}}1 
                {\ -\ }{{\mdseries\ -\ }}3,
}

\lst@AddToHook{EveryLine}{\ifx\lst@language\language@yaml\YAMLkeystyle\fi}
\makeatother

\newcommand\ProcessThreeDashes{\llap{\color{cyan}\mdseries-{-}-}}

\usepackage{microtype}
\usepackage{multirow}

\aclfinalcopy 

\title{SF-QA: Simple and Fair Evaluation Library for Open-domain Question Answering}

\author{
  Xiaopeng Lu$^{2}$\thanks{This work was done during an internship at SOCO}, Kyusong Lee$^{1}$ and Tiancheng Zhao$^{1}$ \\
  SOCO Inc. \\
  $^{1}$\texttt{\{kyusongl,tianchez\}@soco.ai} \\
  Language Technologies Institute, Carnegie Mellon University\\
  $^{2}$\texttt{xiaopen2@andrew.cmu.edu}
  }

\date{}

\begin{document}
\maketitle
\begin{abstract}
Although open-domain question answering (QA) draws great attention in recent years, it requires large amounts of resources for building the full system and it is often difficult to reproduce previous results due to complex configurations. In this paper, we introduce SF-QA: simple and fair evaluation framework for open-domain QA. SF-QA framework modularizes the pipeline open-domain QA system, which makes the task itself easily accessible and reproducible to research groups without enough computing resources. The proposed evaluation framework is publicly available and anyone can contribute to the code and evaluations. 

\end{abstract}

\section{Introduction}
Open-domain Question Answering (QA) is the task of answering open-ended questions by utilizing knowledge from a large body of unstructured texts, such as Wikipedia, world-wide-web and etc. This task is challenging because researchers have to face issues in both scalability and accuracy. In the last few years, rapid progress has been made and the performance of open-domain QA systems has been improved significantly~\cite{chen2017reading,qi2019answering,yang2019end}. Several different approaches were proposed, including two-stage ranker-reader systems~\cite{chen2017reading}, end-to-end models~\cite{seo2019real} and retrieval-free models~\cite{raffel2019exploring}. Despite people's increasing interest in open-domain QA research, there are still two main limitations in current open-domain QA research communities that makes research in this area not easily accessible:

\textbf{The first issue} is the high cost of ranking large knowledge sources. Most of the prior research used Wikipedia dumps as the knowledge source. For example, the English Wikipedia has more than 7 million articles and 100 million sentences. For many researchers, indexing data of this size with a classic search engine (e.g., Apache Lucene~\cite{mccandless2010lucene}) is feasible but becomes impractical when indexing with a neural ranker that requires weeks to index with GPU acceleration and consumes very large memory space for vector search. Therefore, research that innovates in ranking mostly originates from the industry. 

\textbf{The second issue} is about reproducibility. Open-domain QA datasets are collected at different time, making it depends on different versions of Wikipedia as the correct knowledge source. For example, SQuAD~\cite{rajpurkar2016squad} uses the 2016 Wikipedia dump, and Natural Question~\cite{kwiatkowski2019natural} uses 2018 Wikipedia dump. Our experiments found that a system's performance can vary greatly when using the wrong version of Wikipedia. Moreover, indexing the entire Wikipedia with neural methods is expensive, so it is hard for researchers to utilize others' new rankers in their future research. Lastly, the performance of an open-domain QA system depends on many hyperparameters, e.g. the number of passages passed to the reader, fusion strategy, etc., which is another confounding factor to reproduce a system's results.

Thus, this work proposes SF-QA (Simple and Fair Question-Answering), a Python library to solve the above challenges for two-stage QA systems. The key idea of SF-QA is to provide pre-indexed large knowledge sources as public APIs or cached ranking results; a hub of reader models; and a configuration file that can be used to precisely reproduce an open-domain QA system for a task. The pre-indexed knowledge sources enable researchers to build on top of the previously proposed rankers without worrying about the tedious work needed to index the entire Wikipedia. Then the executable configuration file provides a complete snapshot that captures all of the hyperparameters in order to reproduce a result. 

Experiments are conducted to validate the effectiveness of SF-QA. We show that one can easily reproduce previous state-of-the-art open-domain QA results on four QA datasets, namely Open SQuAD, Open Natural Questions, Open CMRC, and Open DRCD. More datasets will be included in the future. Also, we illustrate several use cases of SF-QA, such as efficient reader comparison, reproducible research,  open-source community, and knowledge-empowered applications.

SF-QA is also completely open-sourced~\footnote{\url{ https://github.com/soco-ai/SF-QA.git}}
 and encourages the research community to contribute their rankers or readers into the repository, so that their methods can be used by the rest of the community. 

In short, the contributions of this paper include:
\begin{enumerate}
    \item The proposed open-source SF-QA project that provides a complete pipeline for simplifying open-domain QA research.
    \item A hub of pre-indexed Wikipedia at different years with different ranking algorithms as public APIs or cached results.
    \item Experiments and tutorials that explain use cases and scenarios of SF-QA and validate its effectiveness.
\end{enumerate}


\section{Related Work}
Existing deep learning open-domain QA approaches can be broadly divided into three categories. 

\subsection{Two-stage Approach}
Recent open-domain QA systems mostly use a two-stage ranker-reader approach. Dr.QA~\cite{chen2017reading} uses a modified TF-IDF bag-of-words method as the first-stage retriever. Selected documents are then fed into an RNN-based document reader to extract the final answer span. \citet{wang2018r} leverage reinforcement learning to update both ranker and reader components and shows improvement over Dr.QA in open-domain QA task. \citet{lee2018ranking} focuses on the ranker improvement and uses a learned reranker to boost first stage answer recall. 

Some other works focus on second-stage reader improvement. \citet{yang2019end} adopts a BERT-based reader model \cite{devlin2018bert} instead of the previous RNN-based model and that significantly improved the end-to-end performance. To deal with span extraction in a multi-document setting, \citet{wang2019multi} uses the global normalization approach \cite{clark2017simple} to make the span scores comparable among candidate documents, which improved the performance by a large amount. 

The graph-based ranker-reader approach has also been explored recently. \citet{asai2019learning} proposes a graph-based retriever to retrieve supporting documents recursively based on entity link evidence, and then uses a BERT-based reader model to complete open-domain QA task.


\subsection{End-to-End Approach}
Open-domain QA using the end-to-end approach was not feasible for a long time, because this needs humongous memory to index the corpus and do the vector search. With the emergence of a large pre-trained language model (PLM), researchers revisit this idea and make the end-to-end open-domain QA feasible. \citet{lee2019latent} proposed Open-retrieval QA (ORQA) model, which updates the ranker and reader model in an end-to-end fashion by pre-training the model with an Inverse Cloze Task (ICT). \citet{seo2019real} experiments with considering open-domain QA task as a one-stage problem, and indexing corpus at phrase level directly. This approach shows promising inference speed with compromise in worse performance.


\subsection{Retrieval-free Approach}
Pre-trained language models have got rapid development in recent years. Querying a language model directly to get phrase-level answers becomes a possibility. The T5 model (11B version)~\cite{raffel2019exploring} can reach comparative scores on several open-domain QA datasets, compared with two-stage approaches with far less number of parameters ($\sim$330M). However, as reported in \citet{guu2020realm}, decreasing the number of parameters hurts the model performance drastically. This leaves large room for future research on how to make retrieval-free open-domain QA feasible in the real-world setting.


\section{The Proposed Method}
\begin{figure*}[ht]
\centering
\includegraphics[width=16cm]{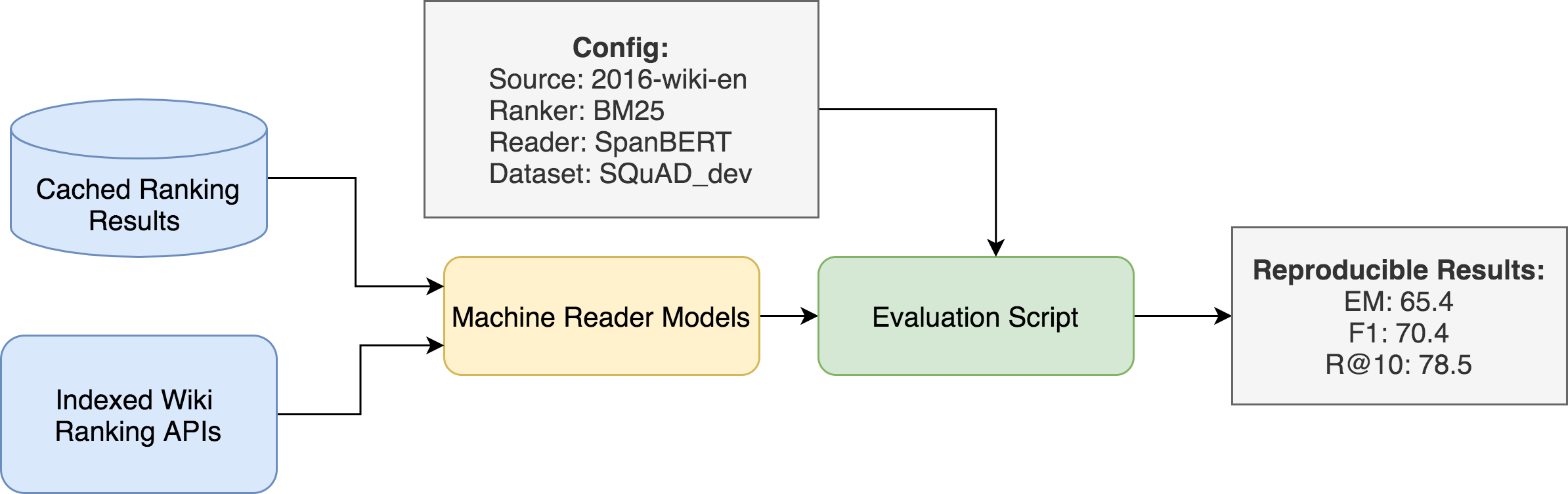}
\caption{Overall pipeline for open-domain QA}
\label{fig:overview}
\end{figure*}

\subsection{Background}
A typical ranker-reader-based open-domain QA system operates as follows: first, a large text knowledge-base is indexed by a ranker, e.g. a full-text search engine. Given a query, the ranker can return a list of relevant passages that may contain the correct answer. How to choose the size of a passage is still an open research question and many choices are available, e.g. paragraph, fixed-size chunks, and sentences. Note that it is not necessary that the ranker needs to return the final passages in one-shot: advanced ranker can iteratively refine the passage list to support multi-hop reasoning~\cite{yang2018hotpotqa,asai2019learning}.

Then given the returned passages, a machine reader model will process all passages jointly and extract potential phrase-level answers from them. A fusion strategy is needed to combine candidate answers and scores from each passage and to create a final list of N-best phrase-level answers by reading these passages. The reason to combine ranker with the reader is to solve the scalability challenge since the state-of-the-art readers are prohibitively slow to process very large corpus in real-time~\cite{chen2017reading,devlin2018bert}.

\subsection{The Proposed Library Overview}
SF-QA is a library that is designed to make it easy to evaluate and reproduce open-domain systems that use ranker-reader architecture. SF-QA decreases the cost of indexing, hosting, and querying large unstructured text knowledge base, e.g. Wikipedia, and also provides a complete configuration snapshot that can be used to replicate a QA system's performance. It is also a place for open-domain QA researchers to share their work, no matter it is innovating in better information retrieval or it is in stronger machine reading comprehension. 

There are four main components in SF-QA: \textit{ranker service}, \textit{reader hub}, \textit{evaluation}, and \textit{pipeline configuration}.

\subsection{Ranker Service}
The goal of the ranker service is to reduce the cost and time to index and query large knowledge source for open-domain QA research using a variety of ranking technologies. Up to date, we have included the BM25~\cite{robertson2009probabilistic} and SPARTA~\cite{zhao2020sparta} ranking methods with several configurations detailed below. More methods will be included and we also welcome community contributions.

Currently, SF-QA supports four ways of document splitting for indexing:
\begin{enumerate}
    \item Sentence: sentence-level indexing
    \item Paragraph: paragraph-level indexing
    \item Chunk: fixed word size indexing
    \item Context: context-level indexing, where the full sentence is always kept, with a maximum number of tokens
\end{enumerate}

Also, Wikipedia dumps at different times are indexed separately so that users can choose to use the same dump as benchmark datasets used. The following versions are included:
\begin{enumerate}
    \item English Wikipedia: 2016/2018/2020
    \item Chinese Wikipedia: 2017/2018/2020 
\end{enumerate}

The returned passage is in the following JSON format: \{$\langle question\_id\rangle$: [{\textit{``score": 42.86,``answer": ``Super Bowl V, the fifth edition of the Super Bowl..."}}, ...]\}, which contains all question ids as key, and top-k retrieved documents and scores as value.

There are two methods to use the ranking results: \textit{cached ranking results} and \textit{ranking API}.

\subsubsection{Cached Ranking Results}
The fastest way to use ranking service for experiments is via cached ranking results. SF-QA provides top-K ranked passages in JSON format for training, validation and test (if publicly available) set. One can directly use the cached results for training or for testing, saving time, and resources for processing the raw data. Another use case is one may use more computationally expensive re-ranking methods to re-rank the top-K passages and then feed them into the reader component. 

\subsubsection{Ranking APIs}
The cached results are very useful for researchers who work on existing datasets and who do not need to have a live system. However, only cached results do not work for new datasets or live QA system that needs to handle user queries. Therefore, SF-QA also provides public API as a service to solve this need. The API is available as a RESTful API and can be reached via HTTPs. Detailed connection documentation can be found on the GitHub.


\subsection{Reader Hub}
Reader hub allows SF-QA's user to specify which reader model to use to extract phrase-level answers. One can either uses their own models by implementing an abstract function or directly load any reader models that are compatible with the Hugging Face Transformer library~\cite{wolf2019huggingface}. SF-QA also includes its own reader model that is optimized for open-domain QA. For example, it offers a BERT reader that is globally normalized~\cite{wang2019multi}, which provides more reliable answer scores to compare multiple candidates' answers from different passages.

Moreover, the reader hub allows the user to define the fusion mechanism that combines the ranking results with reading results. The current implementation supports a linear combination with two free variables, namely the type of score and the weight on reader score. Concretely, the final answer score is computed as follows:
\begin{equation}
    y = (1-\alpha) y_{reader} + \alpha y_{rank}
\end{equation}
where $\alpha$ is a coefficient between 0 and 1. $y_{reader}$ is the reader score, which can either be logits or probability after the softmax layer. $y_{ranker}$ is the ranker score, which depends on the ranking method. One may also specify different normalization strategies to normalize the score from ranker or reader. Normalization strategies include z-normalization, floor normalization etc. Lastly, one may easily add their own strategy by overriding the fusion function.

\subsection{Evaluation}
SF-QA evaluation is designed to offer a multilingual and comprehensive evaluation script that computes the performance of an open-domain QA system and also outputs useful intermediate metrics that are useful for analysis and visualization. For language support, SF-QA evaluation supports English and Chinese. For metrics, it has the most common EM (exact match) and F1 score for the final performance. It also provides other relevant metrics. The following is a list of metrics that are in the output:
\begin{itemize}
    \item Exact match (EM)
    \item F-1 Score
    \item Ranking recall at K 
    \item Oracle ranker score
    \item Mean reciprocal rank (MRR)
\end{itemize}

\subsection{Pipeline Configurations}
\label{sec:config}
The pipeline configuration file is in YAML format, which defines all the hyperparameter for an open-domain QA system to do a forward inference. One can set the configuration for data, ranker ID, and reader ID, fusion strategy and etc. Therefore, the easiest way to share an open-domain QA system for results replication is via providing the right YAML configuration. The following is an example.

\begin{lstlisting}[language=yaml]
# config.yaml
data:
    lang: en
    name: squad
    split: dev-v1.1
ranker:
    use_cached: False
    model:
        name: sparta
        es_index_name: en-wiki-2016
reader:
    model_id: squad-context-spanbert
param:
    n_gpu: 2
    score_weight: 0.8
    top_k: 10

\end{lstlisting}

\section{Use Cases}
SF-QA is designed to be modular and ready to use, with the hope that it can connect people from researchers interested in Question Answering (QA), Information Retrieval (IR), and developers from industries. 
In this section, we illustrate several use cases of SF-QA.

\subsection{Efficient Reader Comparison}
In open-domain QA, the first stage ranker consumes humongous resources in both time, memory, and storage. For researchers without enough computing power, it is not feasible to start open-domain QA research, even if they only want to improve the system on the reader stage. SF-QA provides solutions for researchers with this need. In SF-QA, existing publicly available open-domain QA datasets are already indexed with multiple rankers used in previous open-domain QA research work, currently including BM25~\cite{robertson2009probabilistic}, and SPARTA~\cite{zhao2020sparta}, both with different granularity options. Researchers can call the RESTful API to get the cached ranking results directly if they want to focus on existing open-domain QA datasets, for example, Open SQuAD, Open CMRC, etc. Alternatively, they can call the backend live ranker to get the top retrieved results regarding the input query. We design SF-QA to be completely modular: the researcher is able to pick up a cached ranker and plug in their own reader model to evaluate the open-domain QA results.

\subsection{Reproducible Research}
Reproducibility is another problem that existed in current open-domain QA research. Since the first-stage retriever model needs researchers to collect large-scale data by themselves, it is hard to keep all the settings the same to make fair comparisons. In SF-QA, we collected data following the earliest works' setting~\cite{chen2017reading, yang2019end, kwiatkowski2019natural}. Therefore, researchers can check SF-QA to get data specifications for existing models. Moreover, parameter settings for different models are recorded and saved in another separate configuration file, as shown in the section\ref{sec:config}. Therefore, any existing models in the current SF-QA project can be directly reproduced, which would greatly facilitate researchers in establishing benchmark scores and doing fair comparisons.

\subsection{Knowledge-empowered Applications}
SF-QA framework also considers the needs from an industry perspective. To show the potential of open-domain QA and to encourage more people to join the development of this task, we also provide a RESTful API (with a ready-to-use open-domain QA model in the backend) for users to ask questions and get the phrase-level answers directly as output. We also provide a tutorial to demonstrate that SF-QA can be seamlessly incorporated into RASA \cite{bocklisch2017rasa}, a popular open-source chatbot building platform, with only a few lines of code. We hope that this effort can attract people from different backgrounds to open-domain QA research. 





\section{Experiment Results}

\begin{table}[h]
\centering
\scalebox{0.88}{
\begin{tabular}{p{0.20\textwidth}|p{0.05\textwidth}p{0.04\textwidth}|p{0.05\textwidth}p{0.05\textwidth}} \hline
           & Reported      &              & Reproduced       &             \\
           & EM            & F1           & EM          & F1          \\ \hline
Bertserini~\cite{yang2019end}     & 38.6       &  46.1            & 41.2            &  48.6      \\ 
+DS~\cite{xie2020distant}         & 51.2   &   59.4    &       51.6      &  59.2      \\
Multipassage~\cite{wang2019multi}& 53.0   &   60.9    & 53.2 &  60.7 \\

SpartaQA~\cite{zhao2020sparta}    & 59.3   &  66.5       &      59.3       &    66.5    \\\hline
\end{tabular}}
\caption{Comparison between reported performance and reproduced performance on Open SQuAD.}
\label{tbl:reproduce}
\end{table}

\begin{table*}[ht]
\centering
\begin{tabular}{|l|l|l|l|l|l|}
\hline
& Indexing & Uploading & Retrieval & Reader & Total \\ \hline
\begin{tabular}[c]{@{}l@{}}Traditional\\ Approach\end{tabular} & 16.2 h        & 5.2 h       & 6.1 h         & 4.4 h         & 21.9 h   \\ \hline
SF-QA                                                          & -        & -        & -         & 4.4 h  & 4.4h  \\ \hline
\end{tabular}
\caption{Time elapsed to evaluate open-domain QA using Open SQuAD development set}
\label{tbl:speed}
\end{table*}

\begin{table*}[ht!]
\small
\centering
\begin{tabular}{|l|l|l|l|l|l|l|l|l|l|}
\hline
\multirow{2}{*}{open-domain QA setting} & \multicolumn{3}{l|}{wiki 2016*} & \multicolumn{3}{l|}{wiki 2018}  & \multicolumn{3}{l|}{wiki 2020} \\ \cline{2-10} 
& EM        & F1       & R@1       & EM        & F1       & R@1       & EM        & F1       & R@1       \\ \hline
BM25 + SpanBERT & \textbf{49.2}         & \textbf{56.7}        & \textbf{41.9}              & 45.8         & 53.8        & 39.4       & 41.5         & 49.5        & 35.4             \\ \hline
Sparta + SpanBERT & \textbf{59.3}         & \textbf{66.5}        & \textbf{50.8}                    & 46.5         & 54.4        & 39.3       & 46.4         & 53.9        & 42.2       \\ \hline
\end{tabular}
\caption{Open SQuAD performance using Wikipedia dumps from different years. * represents the dump which SQuAD originally used for annotation.}
\label{tbl:time-compare}
\end{table*}

\subsection{Reproducing Prior Art}
Results in Table \ref{tbl:reproduce} shows the performance comparison between several reported open-domain QA systems and our reproduced results. The first experiment conducted is to reproduce some prior results using SF-QA. We choose Bertserini~\cite{yang2019end}, Bertserini with distant supervision~\cite{xie2020distant}, Multi-passage Bert~\cite{wang2019multi}, and SPARTA~\cite{zhao2020sparta} as three benchmark systems to reproduce. 

To reproduce Bertserini~\cite{yang2019end}, we follow the implementation described in the original paper and first index the 2016 English Wikipedia in paragraph level to get 29.5M documents in total. A BERT-base-cased model is trained with global normalization, following descriptions in the paper. We observe a slight improvement in the open-domain QA result, which may due to the usage of a newer version of the BM25 retriever. The same phenomenon has also been reported in~\cite{xie2020distant}.

For Bertserini with distant supervision~\cite{xie2020distant}, we follow the two-stage distant supervision strategy proposed by the original author, where the model was first fine-tuned using the original SQuAD dataset, and then fine-tuned on the distantly supervised data retrieved from the full Wikipedia. The score we get matches the score reported by the original author.

To reproduce Multi-passage BERT~\cite{wang2019multi}, we first index the Wikipedia corpus using chunk size equals to 100, with a stride of 50 words. A BERT reranker is then trained to rerank the retrieved top 100 documents and the top 30 documents are passed to the reader. In the reader training stage, we train the model using BERT-large-cased model, also with global normalization to make the span score comparable. Our reproduced score matches the score reported in the original paper.

For SpartaQA, we follow the original author's implementation on SPARTA retriever, and index the Wikipedia in the context level with a size of 150. During the reader stage, a SpanBERT ~\cite{joshi2020spanbert} model is used to train the model with distantly supervised data retrieved from Wikipedia with global normalization strategy. The score matches the reported score.

\subsection{Time saved by SF-QA}
This experiment shows results for elapsed time to evaluate open domain question answering with and without the SFQA evaluation framework (Table~\ref{tbl:speed}). Traditionally, we need to build the complete pipeline in order to evaluate the open-domain QA as following steps: (1) \textbf{Indexing}: converting full Wikipedia into sparse or dense representations; (2) \textbf{Uploading:} inserting the text and representations to Elasticsearch (or similar database);  3) \textbf{Retriever}: retrieval n-best candidates from Elasticsearch;  4) \textbf{Reader:} span prediction using machine reading comprehension. We use GeForce RTX 2080 Ti GPU to index the entire Wikipedia dump of the total 89,544,689 sentences. The total amount of elapsed time for open-domain QA is 29 hours without using SF-QA for one experimental setting. In comparison to this, using cached retrieved results provided from SF-QA saves repetitive work in heavy indexing, and it only takes $\sim$ 4 hours to get the final scores.\

\subsection{Model Accuracy v.s. Corpus release year}
We conduct the last experiment to test the robustness of the state-of-the-art system against temporal shifting. Results are reported in Table~\ref{tbl:time-compare}. We observe that model accuracy is largely affected by the version of the Wikipedia dump, showing that it is essential to track the version of the input data and make sure that all open-domain QA researches are reproducible starting from the data input level.


\section{Conclusion}
In conclusion, this paper presents SF-QA, a novel evaluation framework to make open-domain QA research simple and fair. This framework fixes the gap among researchers from different fields, and make the open-domain QA more accessible. We show the robustness of this framework by successfully reproducing several existing models in open-domain QA research. We hope that SF-QA can make the open-domain QA research more accessible and make the evaluation easier. We expect to further improve our framework by including more models in both ranker and reader side, and encourage community contributions to the project as well.

\newpage
\bibliography{eacl2021}
\bibliographystyle{acl_natbib}

\end{document}